\newcommand{\fref}[1]{Fig.~\ref{#1}}
\newcommand{\tref}[1]{Table~\ref{#1}}

\documentclass[final,nopreprintline,11pt]{elsarticle}

\journal{Pattern Recognition}

\usepackage{bbm}
\usepackage{tikz}
\usepackage{pgfplots}
\usepackage{pgfplotstable}
\usepackage{float}
\usepackage{dblfloatfix}
\usepackage{adjustbox}
\usepackage{makecell}
\usepackage{subfig}

\usetikzlibrary{shapes.geometric}
\usepackage{pgfplots}
\pgfplotsset{compat=newest}
\usepackage{amsmath}
\usepackage{amsthm}
\usepackage{booktabs}
\usepackage{amsfonts}

\usepackage{amssymb}

\newcommand{\blue}[1]{#1}
\usepackage[margin=1in]{geometry}
\usepackage[hyphens]{xurl} 
\usepackage[misc]{ifsym}

\begin{document}

\begin{frontmatter}



\title{YOGA: Deep Object Detection in the Wild with Lightweight Feature Learning and Multiscale Attention}

\author{Raja Sunkara}
\ead{rs5cq@mst.edu}
\author{Tie Luo\textsuperscript{\Letter}}
\ead{tluo@mst.edu}


\address{Department of Computer Science\\ Missouri University of Science and Technology, Rolla, MO 65409, USA\vspace{-5mm}}


\begin{abstract}

\blue{We introduce YOGA, a deep learning based yet lightweight object detection model that can operate on low-end edge devices while still achieving competitive accuracy. The YOGA architecture consists of a two-phase feature learning pipeline with a cheap linear transformation, which learns feature maps using only half of the convolution filters required by conventional convolutional neural networks. In addition, it performs multi-scale feature fusion in its neck using an attention mechanism instead of the naive concatenation used by conventional detectors. YOGA is a flexible model that can be easily scaled up or down by several orders of magnitude to fit a broad range of hardware constraints. We evaluate YOGA on COCO-val and COCO-testdev datasets with other over 10 state-of-the-art object detectors. The results show that YOGA strikes the best trade-off between model size and accuracy (up to 22\% increase of AP and 23-34\% reduction of parameters and FLOPs), making it an ideal choice for deployment in the wild on low-end edge devices. This is further affirmed by our hardware implementation and evaluation on NVIDIA Jetson Nano.
\vspace{-1mm}}

\end{abstract}






\end{frontmatter}



\section{Introduction}

Object detection empowered by deep learning has made booming success in diverse applications such as autonomous driving, medical imaging, remote sensing, and face detection. Research in this area has been thriving and the performance competition is fierce. Well-known detectors include the R-CNN series 
\cite{girshick2014rich}, YOLO series 
\cite{redmon2018yolov3}, SSD \cite{liu2016ssd}, RetinaNet~\cite{lin2017focal}, EfficientDet \cite{tan2020efficientdet}, YOLO-Anti~\cite{wang2022yolo}, UDNet~\cite{fang2023udnet}, etc. 
Although the fierce competition has led to better performance in general, it has also resulted in deeper neural network architectures and more complex model designs, implying a need for more training data, more tuning parameters, and longer training and inference time. This would not be suitable for resource-constrained environments such as Internet of Things (IoT) devices at the edge.

Researchers have attempted {\em Pruning and Quantization} methods toward this goal. However, that is an ``aftermath'' approach and the effect is often limited (for example, we applied PyTorch's pruning utility to three popular object detection models and observed a mere improvement of 0\%, 8\%, and -15\%(negative), respectively). To fundamentally address this problem for edge deployment in the wild, a clean-slate design is much more desired.

In this paper, we propose YOGA, a new object detection model based on a resource-conscious design principle. YOGA cuts down model size by up to $34\%$ (cf. Table \ref{tab:testdev}), in terms of number of model parameters and FLOPs, yet notably, achieving competitive accuracy (often even better, by up to $22\%$; cf. Table \ref{tab:testdev}). YOGA consists of (i) a new backbone called CSP\underline{G}hostNet ({\em cross stage partial} GhostNet), (ii) a new neck called \underline{A}FF-PANet ({\em attention feature fusion}-based {\em path aggregation network}), and (iii) a \underline{YO}LO-based head. (The underlined letters account for the coined name, YOGA.) 
Our main idea is twofold. First, to slim down the neural network, we use a two-phase feature learning pipeline with a cheap linear transformation called group convolution throughout the network, which can learn the same number of feature maps as in standard CNNs but using only half of the convolution filters. 
Second, to achieve high accuracy, we 
fuse multi-scale feature maps at the neck using a {\em local attention} mechanism along the channel dimension (besides global attention along the space dimension), rather than using the conventional {\em concatenation} which is essentially equal-weighted. 

Apart from being lightweight and high-performing, YOGA also represents a flexible design in that it can be easily scaled up or down in a wide range by choosing different repetitions of one of its building blocks (CSPGhost; cf. \fref{fig:YOGA}). This makes it easily fit for a broad range of applications with different resource constraints, from small embedded IoT systems to intermediate edge servers and to powerful clouds.

Besides the performance evaluation commonly seen in the literature, we have also implemented YOGA on real hardware, NVIDIA Jetson Nano 2GB (the {\em lowest-end} deep learning device from NVIDIA), and tested its performance to assess its applicability to edge deployment in the wild. \blue{The results are promising (for instance, YOGA-\textit{n} runs at 0.57 sec per 640x640 image, which is close to real-time) and will surely be much more responsive on less-restrictive hardware (e.g., Jetson Nano 4GB, or Jetson TX2).}

In summary, the contributions of this paper are:
\begin{itemize}
\item We propose YOGA, a new object detection model 
that learns richer representation (via attention based multi-scale feature fusion) with a much lighter model (reducing nearly half convolution filters via group convolution).

\item We provide a theoretical explanation of how {\em label smoothing} facilitates backpropagation during training, by mathematically analyzing how the loss gradient vector is involved in the recursive backpropagation algorithm when label smoothing is used. We also overcome a GhostNet overfitting issue using a hyper-parameter tuning method based on Genetic Algorithm.

\item We compare YOGA with a large variety of (over 10) state-of-the-art deep learning object detectors (as YOGA can be easily scaled up or down so we can make fair comparison with models at different levels of scales). The results demonstrate the superiority of YOGA on the joint performance of model size and accuracy. 

\item We also migrate YOGA to real hardware to assess its usability in the wild. Our experiments show that YOGA is well suited for even the lowest-end deep learning edge devices.

\end{itemize}

\section{Related Work and Preliminaries}\label{sec:relwk}

Current state-of-the-art object detection models are convolutional neural network (CNN) based and can be categorized into one-stage and two-stage detectors, or anchor-based or anchor-free detectors. A two-stage detectors first use region proposals to generate coarse object proposals, and then use a dedicated per-region head to classify and refine the proposals. In contrast, one-stage detectors skip the region proposal step and run detection directly over a dense sampling of locations. Anchor-based methods use {\em anchor boxes}, which are a predefined collection of boxes that match the widths and heights of training data objects, to improve the loss convergence during training. We provide a classification of some well-known object detection models in Table \ref{table:cls}. For a detailed review of such methods, the reader is referred to a comprehensive survey~\cite{liu2020deep}. For a overview of deep-learning based methods for salient object detection in videos, refer to \cite{wang2020overview}.

\begin{table}[ht]
\caption{A taxonomy of object detection models.}
\centering
\resizebox{.75\linewidth}{!}{%
    \begin{tabular}{|c|c|c|}
     \toprule
     
     {Model} & {Anchor-based} & {Anchor-free} \\ \midrule
     
     
     One-stage & \makecell{
Faster R-CNN~\cite{ren2015faster}, SSD~\cite{liu2016ssd},\\
RetinaNet~\cite{lin2017focal}, EfficientDet~\cite{tan2020efficientdet}, \\
YOLO ~\cite{redmon2018yolov3}
} & 
     \makecell{FCOS~\cite{tian2019fcos}, CenterNet~\cite{duan2019centernet},\\
    DETR~\cite{carion2020end}, YOLOX~\cite{ge2021yolox}} \\ \midrule

     Two-stage & \makecell{R-CNN~\cite{girshick2014rich},\\
     Fast R-CNN~\cite{girshick2015fast}\\
      }
     & \makecell{  RepPoints~\cite{yang2019reppoints}, \\
     CenterNet2
     ~\cite{duan2019centernet}}\\

     \bottomrule
    \end{tabular}

\label{table:cls}
}
\end{table}

Generally, one-stage detectors are faster than two-stage ones and anchor-based models are more accurate than anchor-free ones. Thus, in the YOGA design, we focus on the one-stage and anchor-based models, i.e, the first cell of Table \ref{table:cls}.

\begin{figure}[htb]
    \centering
    \includegraphics[trim=8mm 0 0 2mm,clip,width=0.75\linewidth]{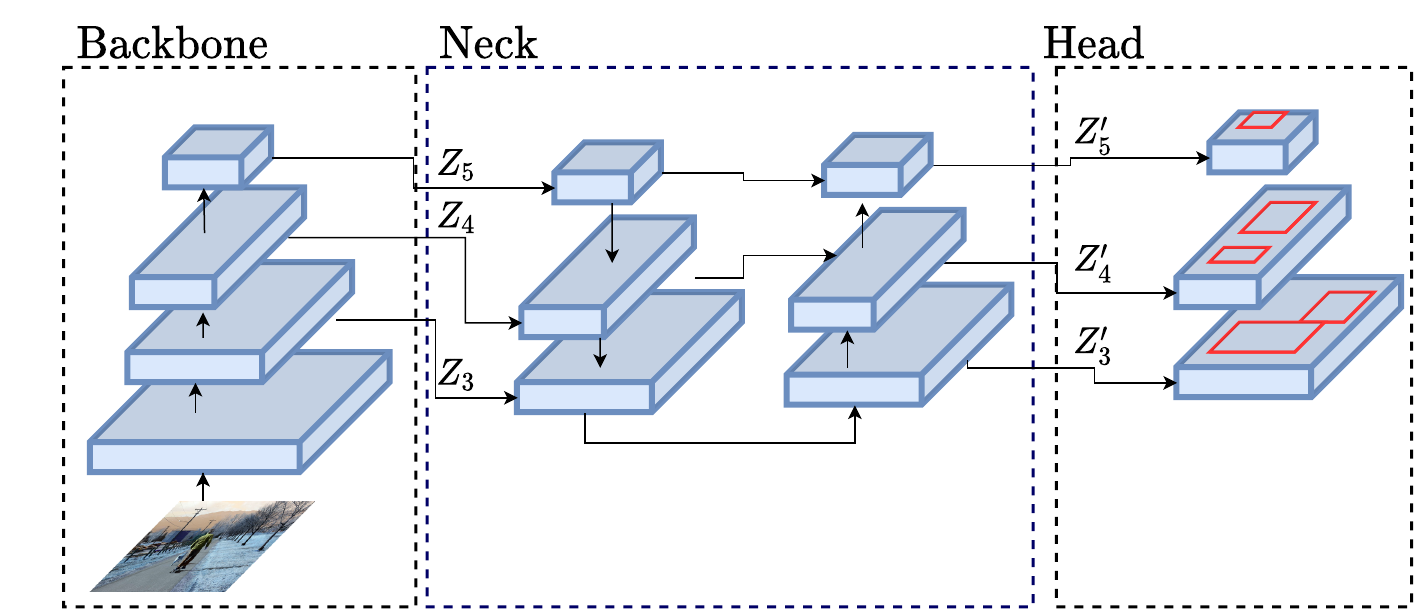}
    \caption{A one-stage object detection model generally consists of a {\em backbone} for feature extraction, a {\em neck} for feature fusion, and a {\em head} for regression and classification.}
    \label{fig:object_de}
\end{figure}

A typical one-stage object detection model is depicted in Fig.~\ref{fig:object_de}. It consists of a CNN-based {\em backbone} for visual feature extraction and a detection {\em head} for predicting class and bounding box of each contained object. In between, a {\em neck} is added to combine features at multiple scales to produce semantically strong features for detecting objects of different sizes.

\section{Design of YOGA}

\begin{figure*}
    \centering
    \includegraphics[width=\linewidth]{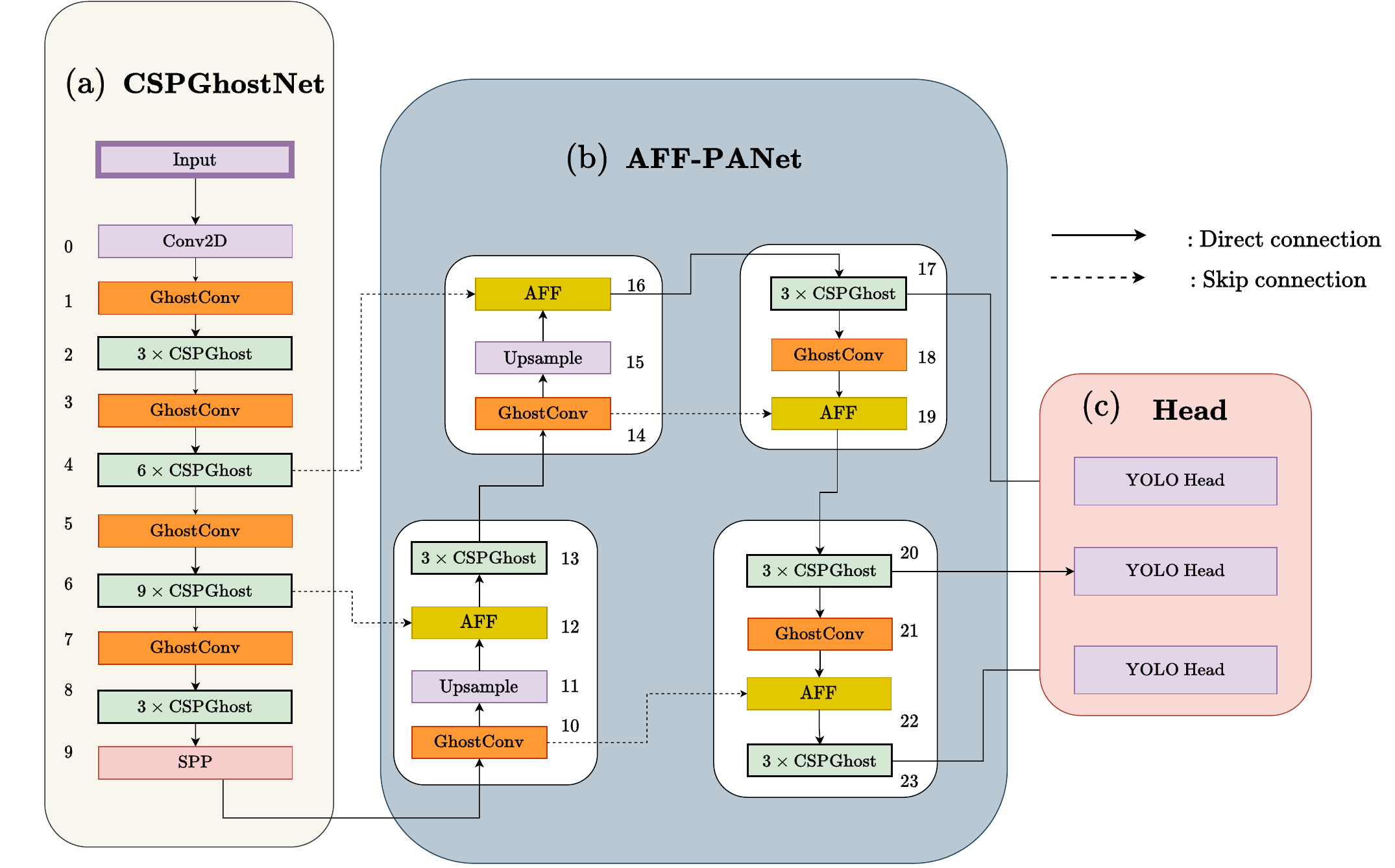}
    \caption{The YOGA architecture: (a) Backbone, (b) Neck, (c) Head. Zoom-in view of CSPGhost module (light green) is provided in \fref{fig:YOGA2}. The $n \times$ repetition allows our model to scale up and down easily.}
    \label{fig:YOGA}
\end{figure*}

An overview of the YOGA architecture is given in Fig.~\ref{fig:YOGA}.

\begin{figure*}
    \centering
    \includegraphics[width=0.85\linewidth]{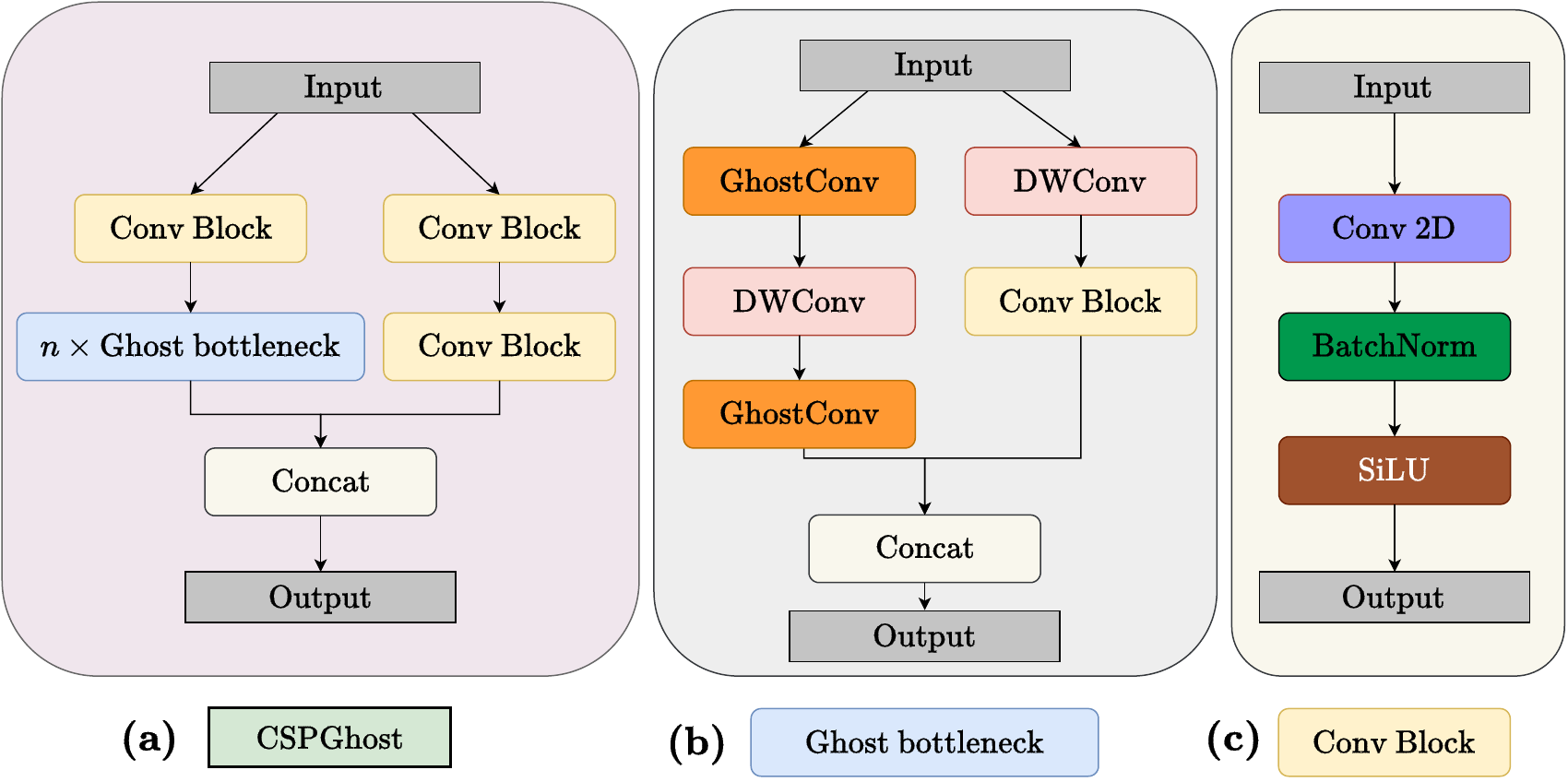}

    \caption{(a) Internal structure of our CSPGhost module. (b) The Ghost bottleneck layer (light blue in CSPGhost), where DWConv stands for depth-wise convolution. (c) The Conv Block (yellow) used in both (a) and (b).}
    \label{fig:YOGA2}
\end{figure*}

\subsection{Backbone: CSPGhostNet}\label{sec:backbone}
 
Our design of backbone, called CSPGhostNet, is motivated by two observations and guided by two corresponding aims. First, we identify that standard CNNs add many redundant features in order to learn better representation of input images. For example, ResNet~\cite{he2016deep} generates numerous similar feature maps. Such designs come at a price of high computational cost and heavyweight models. Therefore, we pose the following question as our first aim: Is it possible to generate the same number of features with similar (necessary) redundancy but using much less computation and less parameters? 

Second, we observe that the training and inference time of current deep learning models has a large room to improve. We are therefore motivated to also speed up training and inference processes in our context, object detection.

To achieve the first aim, we adapt GhostNet~\cite{han2020ghostnet} to exploit a low-cost two-phase convolutional pipeline. For the second aim, we integrate half of the feature maps across backbone and neck from the beginning to the end to create a shortcut, by leveraging CSPNet~\cite{wang2020cspnet}. 
Specifically, it splits each input feature map into two parts, feeds one part through a group of convolution blocks while letting the other part bypass those blocks, and merges these two branches at the end via concatenation. This shortcut also reduces the repetition of gradient information during backpropagation. In the following, we focus on explaining how we achieve the first aim using GhostNet because CSPNet can be applied without much change to its original architecture (however, we are the first that creates a new module combining GhostNet and CSPNet).

\blue{Ghost bottleneck (G-bneck) as in \fref{fig:YOGA2}(b), which we draw {\em based on} the GhostNet paper \cite{han2020ghostnet} (but does not exist in \cite{han2020ghostnet}), was designed specially for small CNNs. It is not trivial to use G-bneck to build medium and large CNNs. In fact, this is also the reason why GhostNet, which is built on top of G-bneck, was compared with only small neural nets like MobileNetv2 and MobileNetv3. To overcome this limit, we designed a new CSPGhost module as in \fref{fig:YOGA2}(a), where G-bneck layer is just part of it (the light blue block). This CSPGhost module allows us to build medium and large CNNs.}

CSPGhost (in light green) is located at multiple positions in both our backbone and neck (see \fref{fig:YOGA}), and its internal structure is shown in~\fref{fig:YOGA2}(a). CSPGhost contains a Ghost bottleneck layer (in light blue) and multiple Conv Blocks (in light yellow). Each Conv Block consists of a 2D Convolution, a BatchNorm, and a SiLU non-linear activation function (\fref{fig:YOGA2}(c)). The Ghost bottleneck layer is similar to ResNet's basic residual block that integrates several 
convolutional layers and short-cut connections. It mainly consists of two GhostConv modules (in orange) with depth-wise convolution in between: the first GhostConv module acts as an expansion layer that increases the number of channels, while the second GhostConv module reduces the number of channels to match the input shortcut path, after which the input of the first GhostConv and the output of the second GhostConv is connected by the shortcut through the depth-wise convolution and Conv block.



\begin{figure}[t]
    \centering
    \includegraphics[width=.8\linewidth,clip,trim=0 5mm 7mm 0]{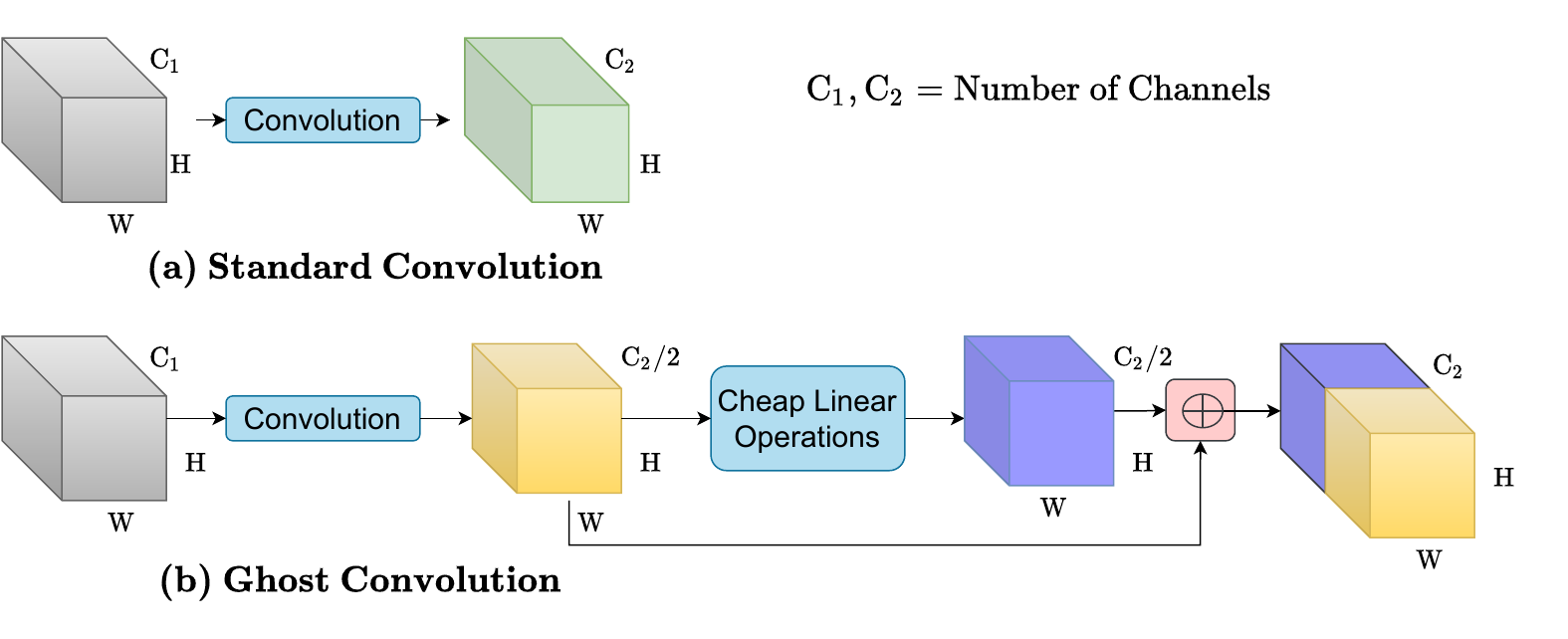}
    \caption{Ghost convolution vs. standard convolution.}
    \label{fig:ghost}
\end{figure}

The GhostConv block (in orange) stands for Ghost convolution and its structure is given in Fig.~\ref{fig:ghost}. In standard convolution, $C_2$ filters each of depth $C_1$ will be used to transform an input feature map of depth $C_1$ to an output feature map of depth $C_2$, as shown in Fig.~\ref{fig:ghost}a. In contrast, GhostConv uses only $C_2/2$ standard filters to generate an intermediate feature map, denoted by $X_a$, of depth $C_2/2$, and then applies a {\em group convolution} with $C_2/2$ groups to $X_a$ to generate a feature map, $X_b$, of identical depth $C_2/2$. Group convolution with $C_2/2$ groups is a cheap linear transformation which only performs {\em per-channel} instead of cross-channel convolution as in the standard convolution. Finally, the two feature maps $X_a$ and $X_b$ are concatenated to obtain the output feature map, which has a depth of $C_2$.

Mathematically, we can formulate this process as follows:
\begin{align}
X_a &= \boldsymbol{w_1} \otimes \boldsymbol{x_0} \text{ (first half; std conv)}\nonumber \\
X_b &= \boldsymbol{w_2} \; \tilde{\otimes} \; X_a \text{ (second half; group conv)}\\
\boldsymbol{y} &=  X_a \oplus X_b \text{ (output feature maps)} \nonumber
\end{align}
where $\otimes$ denotes the standard convolution, $\tilde{\otimes}$ denotes the group convolution, and $\oplus$ denotes concatenation along the channel dimension. Thus, we can see that GhostConv adopts ordinary convolution to generate a few intrinsic feature maps and then utilizes cheap linear operations to augment the features and increase the number of channels.

Because of this, GhostConv can speed up the convolution process as well as reduce the number of parameters by 2 approximately. In general, the improvement factor is $s = C_2/D(X_a)$ where $D(X)$ is the number of channels of a feature map $X$. Based on the empirical analysis in \cite{han2020ghostnet}, $s=2$ results in the best performance. Therefore, we choose half filters to generate intermediate feature maps in GhostConv as shown in \fref{fig:ghost}. 

Our redesigned backbone CSPGhostNet enables YOGA to substantially reduce the number of parameters and FLOPs 
without sacrificing its detection performance (mAP). Moreover, as a general guideline in deep learning, less parameters also tend to imply a more generalizable neural network.

Finally, we add spatial pyramid pooling (SPP)~\cite{he2015spatial} to the tail of our backbone network in order to increase the receptive field. 

\subsection{Neck: AFF-PANet}

We also design a new neck architecture called AFF-PANet that addresses a fundamental problem in object detection: {\em feature fusion}. An object detection task inevitably requires fusing low-level and high-level feature maps extracted from the backbone. However, current research all centers around fusing these feature maps using a naive concatenation with no learning involved. As illustrated in \fref{fig:object_de}, such a neck simply stacks the feature maps $Z_3$, $Z_4$ and $Z_5$ along the channel dimension, and then applies a standard convolution to match the output channels.

The problem with this kind of naive fusion is that {\em concatenation essentially treats each feature map equally}, but the features learned by the backbone have multiple scales and larger-scale ones tend to overshadow smaller-scale ones. Therefore, we propose to incorporate {\em learning} into the fusion process using an {\em attention mechanism}. However, this is non-trivial because typical attention methods such as SENet~\cite{hu2018squeeze} cannot be directly applied to multi-scale features. The underlying reason is that those channel attention mechanisms use an extreme and coarse feature descriptor that implicitly assumes that large objects occupy a large portion of space and averages the feature maps across the spatial dimension, which would wipe out much of the image signal present in small objects. More specifically, such methods compress each feature map into a scalar and thus the average of feature maps along the spatial dimension becomes very small, resulting in poor detection of small objects. In fact, these are {\em global attention} mechanisms alone which cannot well handle multi-scale feature fusion.

Our design is the first that introduces AFF~\cite{dai2021attentional} into the area of object detection in order to add {\em local attention} to feature fusion (besides global attention), and the first that incorporates it in PANet~\cite{liu2018path} to shorten the pathway of passing feature information to the head. \blue{The paper \cite{dai2021attentional} proposed an AFF module in Feature Pyramid Networks (FPN), but in our case, we integrated the AFF module into the Path Aggregation Network (PANet) and build a new neck architecture called AFF-PANet, which we explain the details below.}

\begin{figure}[t]
    \centering
    \includegraphics[width=0.9\linewidth]{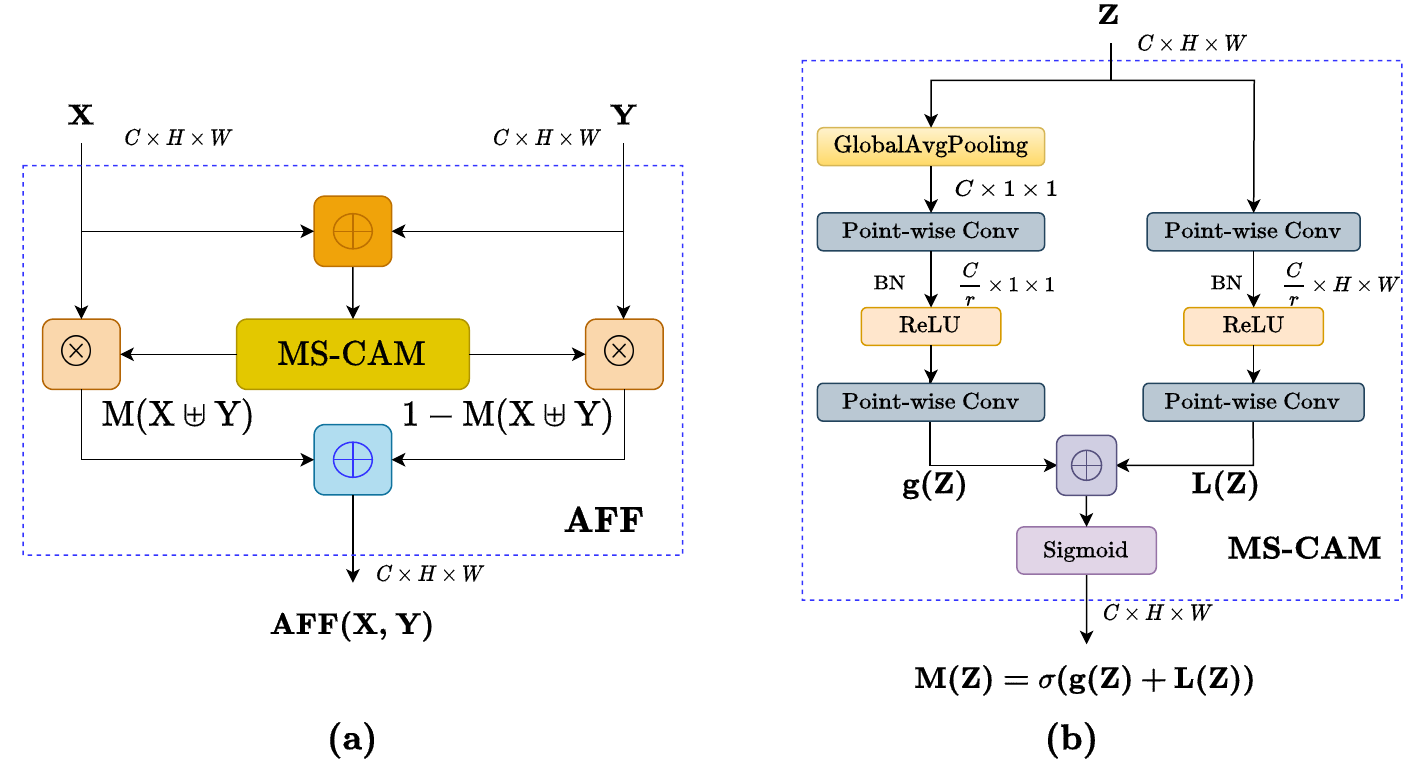}
    \caption{(a) Attention Feature Fusion (AFF). (b) Multi-scale channel attention module (MS-CAM).}
    \label{fig:aff}
\end{figure}

AFF uses a multi-scale channel attention module (MS-CAM) as depicted in \fref{fig:aff}. 
Given an intermediate feature map $\mathbf{Z} \in \mathbb{R} ^{C \times H \times W}$, 
there are two network pathways, one computing a global channel context $\mathbf{g(Z)}$ and the other is responsible for computing a local channel context $\mathbf{L(Z)}$. The two contexts $\mathbf{g(Z)}$ and $\mathbf{L(Z)}$ are then combined through a broadcasting addition operation, followed by a Sigmoid non-linearlity to map values into the range of 0-1, to obtain the attentional weights $\mathbf{M(Z)} = \sigma(\mathbf{g(Z)}\oplus \mathbf{L(Z)})$. Therefore, by ushering AFF into object detection, we mainly exploit its local attention pathway, i.e., $\mathbf{L(Z)}$. 

Mathematically, this process can be expressed as
\begin{align}
    \mathbf{AFF(X, Y)} = \mathbf{M(X} \uplus \mathbf{Y)} \odot \mathbf{X} + (1 - \mathbf{M(X} \uplus \mathbf{Y)}) \odot \nonumber \mathbf{Y} 
\end{align}
where $\mathbf{X} \in \mathbb{R}^{C \times H \times W}$ is a low-level feature map, and $\mathbf{Y} \in \mathbb{R}^{C \times H \times W}$ is a high-level feature map, $\uplus$ denotes the initial feature integration which we choose to be element-wise summation, $\mathbf{M(X} \uplus \mathbf{Y)}$ is a MS-CAM function that computes fusion weights between 0 and 1, which are finally applied to $\mathbf{X}$ and $\mathbf{Y}$ to form a weighted average. 
Corresponding to \fref{fig:YOGA}, $\mathbf{X}$ and $\mathbf{Y}$ are the outputs of blocks 4 and 15 respectively, and $\mathbf{AFF(X, Y)} \in \mathbb{R}^{C \times H \times W}$ is the fused feature as the output of block 16.

We also incorporate PANet in our neck. The reason is as follows. The feature pyramid network (FPN)~\cite{lin2017fpn} is comprised of only a top-down pathway to concatenate feature maps (cf. \fref{fig:object_de} left part of neck), but PANet adds a bottom-up pathway and multiple lateral connections. This addition will help shorten the path of passing feature information from earlier layers to the head through only a few convolutional layers, thereby learning richer representations for multi-scale objects (as shorter paths have a better gradient flow from earlier CNN layers to the neck).

In summary, our AFF-based neck design introduces learning into multi-scale feature fusion by combining feature maps using learnable weights rather than naive concatenation.

\subsection{Head: YOLO}

The purpose of the head is to perform dense predictions, where each prediction consists of an object confidence score, a probability distribution of the object classes, and bounding box coordinates. A head makes these predictions based on the feature maps ($Z^{\prime}_{3}$, $Z^{\prime}_{4}$, and $Z^{\prime}_{5}$ as shown in \fref{fig:object_de}), obtained from the neck. 

We adopt the YOLO head architecture which consists of a $3 \times 3$ convolution layer followed by a $1 \times 1$ convolution layer. The number of filters used in this $1 \times 1$ convolution layer is $N (C + 5)$, where $C$ is the number of classes and $N$ is the number of anchor boxes (each prediction is made by using an anchor at one of three different scales). The output of the head is post-processed by non-maximum suppression (NMS) to eliminate redundant and low-confidence bounding boxes.

\subsection{Label Smoothing}

We use a regularization technique called label smoothing \cite{szegedy2016rethinking} to improve backpropagation gradients during neural network training. Unlike one-hot vector where the entire probability mass is concentrated on a single true class, label smoothing weighs $1-(K-1)\epsilon$ on the true class and $\epsilon$ on the remaining $K-1$ classes. This section provides an in-depth mathematical explanation of how this method helps backpropagation during model training, as \cite{szegedy2016rethinking} proposed it only heuristically.

Given an input sample, let $\mathbf{y}$ be its true label encoded by label smoothing, and $\mathbf{y}_n$ be its prediction made at a neural network's last (the $n$-th) layer. Using the cross-entropy loss $L(\mathbf{y},\mathbf{y}_n) = - \sum_{i=1}^{K} y_{i} \log y_{n_i}$, the gradient of this loss with respect to the predicted output $\mathbf{y}_n$ is given by
\begin{align}
\nabla_{\mathbf{y}_n}L = 
      \begin{cases}
      -\frac{1-(K-1)\epsilon}{y_{n_c}}, & \text{for the true class $c$} \\ 
       -\frac{\epsilon}{y_{n_i}}, & \text{any other classes $i$} \end{cases}
\label{form:label_smoothing}
\end{align}
This gradient is then applied to the recursive backpropgation step given below:
\begin{align}
\begin{split}
&\nabla_{\mathbf{z}_k}L =  (\nabla_{\mathbf{y}_k}L) J_{\mathbf{y}_k}(\mathbf{z}_k) , \,\,\, \nabla_{\mathbf{y}_{k-1}}L = (\nabla_{\mathbf{z}_k}L) \mathbf{W}_{k} \\
&\nabla_{\mathbf{W}_k}L =  \mathbf{y}_{k-1} \nabla_{\mathbf{z}_k}L, \qquad \nabla_{\mathbf{b}_k}L =  \nabla_{\mathbf{z}_k}L 
\end{split}
\label{form:backprop}
\end{align} 
where $\mathbf{z}_k$ and $\mathbf{y}_k$ represent the pre-activation and post-activation vectors, respectively, at layer $k$. We compute Jacobian $J_{\mathbf{y}_n}(\mathbf{z}_n)$ using the last layer activation function, and then apply the recursion \eqref{form:backprop} from $k = n$ (last layer) to 1 (first layer) to compute all the gradients. 
The Jacobain matrix $J_{\mathbf{y}_k}(\mathbf{z}_k)$ and weight matrix $\mathbf{W}_k$ for the $k$-th layer are given, respectively, by 
\[
\left[
\begin{matrix}
\frac{\partial y_{k_1}}{\partial z_{k_1}} & \frac{\partial y_{k_1}}{\partial z_{k_2}} & \hdots & \frac{\partial y_{k_1}}{\partial z_{k_D}}\\
\frac{\partial y_{k_2}}{\partial z_{k_1}} & \frac{\partial y_{k_2}}{\partial z_{k_2}} & \hdots & \frac{\partial y_{k_2}}{\partial z_{k_D}}\\
\vdots & \vdots & \ddots & \vdots\\
\frac{\partial y_{k_M}}{\partial z_{k_1}} & \frac{\partial y_{k_M}}{\partial z_{k_2}} & \hdots & \frac{\partial y_{k_M}}{\partial z_{k_D}}\\
\end{matrix}
 \right],
\left[
\begin{matrix}
w^{(k)}_{11} &w^{(k)}_{21} &\hdots & w^{(k)}_{D_{k-1}1}\\
w^{(k)}_{12} &w^{(k)}_{22} &\hdots & w^{(k)}_{D_{k-1}2}\\
\vdots & \vdots & \ddots & \vdots\\
w^{(k)}_{1D_k} &w^{(k)}_{2D_k}& \hdots & w^{(k)}_{D_{k-1}D_k}\\
\end{matrix}
\right]
\]
where $D_{k}$ represents the number of neurons in the $k$-th layer.

For the label-smoothing based loss, all the entries of the gradient $\nabla_{\mathbf{y}_n}L$ (Eq.~\ref{form:label_smoothing}) are non-zero. As this gradient vector is multiplied with the Jacobian $J_\textbf{y}(\textbf{z})$ and the weight matrix $\mathbf{W}_k$ in the recursive backpropagation step, using such gradient to update weights during gradient descent would significantly mitigate the gradient vanishing problem and thus help the training and convergence of deep neural networks.

\section{Performance Evaluation}

For an extensive performance evaluation, we compare YOGA with a large number of state-of-the-art object detection models as our baselines, including YOLOv5, EfficientDet, YOLOX, YOLOv4, PP-YOLO, DETR, Faster-RCNN, SSD512, etc. and the complete list can be seen from Tables \ref{tab:validation_table} and \ref{tab:testdev}. \blue{(All our results are fully reproducible, and our code will be open-sourced upon acceptance.)} 

\textbf{Model scaling.} As mentioned before, YOGA can easily scale up or down to suit different application or hardware needs. For example, we have tested that its Nano version YOGA-\textit{n} can run near real-time on Jetson Nano and its Large version YOGA-\textit{l} can run real-time on a V-100 GPU.

Specifically, one can scale YOGA by simply adjusting the number of filters in each convolutional layer (i.e., {\em width} scaling) and the number of convolution layers in the backbone (i.e., {\em depth} scaling) to obtain different versions of YOGA, such as Nano, Small, Medium, and Large. The width and depth scaling will result in a new width of $\lceil n_w \times \text{width factor} \rceil_8$ and a new depth of $\lceil n_d \times \text{depth factor} \rceil$, respectively, where $n_w$ is the original width and $n_d$ is the original number of repeated blocks (e.g., 9 as in $9 \times$ CSPGhost as in Fig. \ref{fig:YOGA}), and $\lceil \cdot \rceil_8$ means rounded off to the nearest multiple of 8. The width/depth factors are given in Table \ref{tab:scaling}, where YOGA-\textit{n}/\textit{s}/\textit{m}/\textit{l} correspond to the Nano, Small, Medium, and Large versions of YOGA, respectively.


\begin{table}[h]
\caption{Depth and Width scaling factors in YOGA.}
    \centering
\resizebox{.45\linewidth}{!}{%
    \begin{tabular}{lll}
     \hline
     
     {Model} & {Depth Factor} & {Width Factor}  \\ \midrule
     
     YOGA-\textit{n} & 0.50  & 0.33 \\
     YOGA-\textit{s} & 0.57  & 0.68 \\
     YOGA-\textit{m} & 1.00 & 1.20 \\
     YOGA-\textit{l} & 1.50  & 1.41 \\
     \bottomrule
    \end{tabular}
    }
    \label{tab:scaling}
\end{table}

\subsection{Experiment setup}

\textbf{Dataset and metrics.} We use the COCO-2017 dataset~\cite{lin2014microsoft} which is divided into {\tt train2017} (118,287 images) for training, {\tt val2017} (5,000 images; also called {\tt minival}) for validation, and {\tt test2017} (40,670 images) for testing. We use a wide range of state-of-the-art baseline models as listed in Tables \ref{tab:validation_table} and \ref{tab:testdev}. We report the standard metric of average precision (AP) on {\tt val2017} under different IoU thresholds [0.5:0.95] and object sizes (small, medium, large). We also report the AP metrics on {\tt test-dev2017} (20,288 images) which is a subset of {\tt test2017} with accessible labels. However, the labels are not publicly released but one needs to submit all the {\em predicted} labels in JSON files to the {\tt CodaLab COCO Detection Challenge} \cite{codalab} to retrieve the evaluated metrics, which we did.


\textbf{Training.} We train different versions (nano, small, medium, and large) of YOGA on {\tt train2017}. Unlike most other studies, we {\em train from scratch without using transfer learning}. This is because we want to examine the {\em true learning capability} of each model without being disguised by the rich feature representation it inherits via transfer learning from ideal (high quality) datasets such as ImageNet.   This was carried out on our own models (YOGA-n/s/m/l) and all the existing YOLO-series models (v5, X, v4, and their scaled versions like nano, small, large, etc.). The other baseline models still used transfer learning because of our lack of resource (training from scratch consumes an enormous amount of GPU time). However, note that this simply means that {\em those baselines are placed in a much more advantageous position} than our own models as they benefit from high quality datasets.

We apply this the same way to our YOLO series baselines (v5, X, v4, and their scaled versions) as well. On the other hand, other baseline models still use (and thus benefit from) transfer learning because of lack of resource (training from scratch consumes enormous time and GPU resources). However, this simply means that those baselines are in a {\em much more advantageous} position than YOGA.

\begin{figure}
    \centering
    \includegraphics[width = 0.7\textwidth]{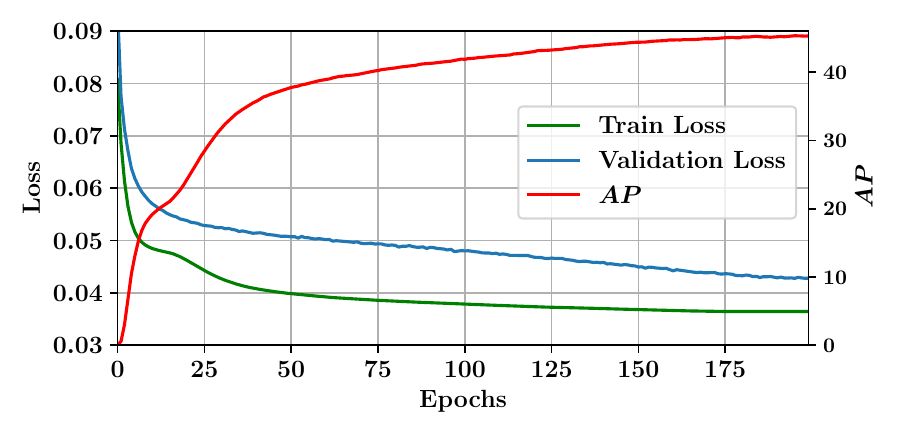}
    \caption{A training history plot. The training loss (green line) and validation loss (blue line) refer to the localization loss. The red line denotes AP values on validation data.}
    \label{figure:training}
\end{figure}

\textbf{Hyperparameter Tuning.} We use the Genetic Algorithm (GA) to tune hyperparameters for YOGA. We ran GA on $m=20$ hyperparameters for 200 generations on a subset of the COCO dataset. Our choice of $m=20$ is based on the Vapnik–Chervonenkis (VC) inequality:
\begin{align}
P[|E_{in} - E_{out}| > \epsilon] \leq 4 m_{h}(2N) e^{-\frac{1}{8} \epsilon ^2 N}
\label{equation:vc}
\end{align}
where $E_{in}$ is the error on the validation set, $E_{out}$ is the error on the test set, $N$ is the number of validation samples, and $m_h$ is the growth function of a hypothesis set defined by $m$. For a small $\epsilon = 0.05$ and with probability 95\%, we choose $N \ge 10\times$ VC-dimension by rule of thumb for good generalization ($E_{in} \approx E_{out}$), where VC-dimension~\cite{abu1993hints} is the number of independent parameters which is upper-bounded by $m$. Since COCO validation data contains $N=5000$ images, choosing $m=20$ satisfies the above inequality, which ensures the difference between test and validation error to be arbitrarily small according to \eqref{equation:vc}. Therefore, our mAP estimates computed from validation data will be reliable to use and is a good proxy for mAP on test data.

\blue{The hyperparameters of GA-based optimization are as follows. It uses the SGD optimizer with momentum 0.937, weight decay of 0.005, a learning rate that linearly increases from 0.0033 to 0.01 for the first three epochs, and then decreases using the Cosine decay strategy to a final value of 0.001. The total number of epochs is 200, which is chosen based on our observation as shown in Fig. 6, where the model enters the overfitting region beyond 200 epochs.}

\blue{In neural network training, a larger batch size can lead to faster training, but it also requires more memory. On the other hand, a smaller batch size may require more training steps, but it may also be more memory efficient. When training on a GPU, the available memory is a limiting factor on the maximum batch size. Therefore, we trained our YOGA nano and small models on 4 V-100 32 GB GPU with a batch size of 128, and medium and large models with batch size 32. We employed CIoU loss for objectness and cross-entropy loss for classification. To mitigate overfitting, we applied several data augmentation techniques following YOLOv5, 
  including photometric distortions of hue, saturation, and value, as well as geometric distortions such as translation, scaling, shearing, fliplr and flipup. Multi-image enhancement techniques such as mosaic and cutmix were also employed.}


For baselines, we use their best hyperparameters stated in their respective papers, or given in their respective online repositories.

\begin{table}
 \caption{Results on {\tt MS-COCO \bf Validation} Dataset. Percentages are in comparison against the {\bf closest} performer.}
\centering
\resizebox{\linewidth}{!}{%
    
    \begin{tabular}{llllllll}

     \hline
     \hline

     {Models} & {Backbone} & {Image-size} & {AP} &{AP\textsubscript{S}} & {Params (M)} & {FLOPs (B)}  \\ \midrule

     
     \textbf{YOGA-\textit{n}}& CSPGhostNet-\textit{n}& $640\times 640$&\textbf{32.3} (\textcolor{blue}{\textbf{+15.35\%}}) & \textbf{15.2}(\textcolor{blue}{\textbf{+7.4\%}})& 1.9  & 4.9  \\
     YOLO-\textit{n}  \cite{yolov5}& - & $640 \times 640$& 28.0 & 14.14& 1.9  & 4.5   \\
     YOLOX-Nano \cite{ge2021yolox}& - & $640\times640$&25.3 &- & 0.9  & 1.08   \\ \midrule


     \textbf{YOGA-\textit{s}}& CSPGhostNet-\textit{s} & $640\times640$&\textbf{40.7} (\textcolor{blue}{\textbf{+8.8\%}}) & \textbf{23.0}(\textcolor{blue}{\textbf{+9.0\%}})& 7.6 (\textcolor{red}{\text{+5\%}})   & 16.6 (\textcolor{red}{\text{+0.6\%}})  \\
     YOLO-\textit{s} \cite{yolov5} &- & $640\times640$& 37.4 &21.09 & 7.2  & 16.5   \\
     YOLOX-S \cite{ge2021yolox} &- & $640\times640$& 39.6 & - & 9.0  & 26.8   \\ 
     YOLOv7-tiny \cite{wang2022yolov7}&-&$640\times640$ & 38.7 & - & 6.2 & 13.8 \\ \midrule
     \textbf{YOGA-\textit{m}}& CSPGhostNet-\textit{m} & $640\times640$& 45.2 & 28.0 &\textbf{16.3}  (\textcolor{blue}{\textbf{-23\%}}) & \textbf{34.6} (\textcolor{blue}{\textbf{-29\%}})   \\
     YOLO-\textit{m} \cite{yolov5} & - & $ 640\times640$& 45.4 & 27.9 & 21.2  & 49.0   \\
     YOLOX-M \cite{ge2021yolox} & - &$ 640\times640$& 46.4 & -& 25.3  & 73.8    \\ \midrule


     \textbf{YOGA-\textit{l}}& CSPGhostNet-\textit{l} & $640\times640$& 48.9 &31.8& \textbf{33.6}  (\textcolor{blue}{\textbf{-27.7\%}}) &\textbf{71.8} (\textcolor{blue}{\textbf{-34\%}})  \\
     YOLO-\textit{l} \cite{yolov5} & - & $ 640\times640$& 49 &31.8& 46.5  & 109.1    \\
     YOLOX-L \cite{ge2021yolox} & - & $640\times640$& 50.0 &-& 54.2  & 155.6   \\ 

     YOLOv7 \cite{wang2022yolov7} &-&$640\times640$ & 51.2 & - & 36.9 & 104.7 \\
     
     \bottomrule

     
     
     HTC \cite{wang2020deep} & HRNetV2p-W48 & $800 \times 1333$ & 47.0 &28.8 & 79.42 &399.12  \\
     
     HoughNet \cite{samet2020houghnet} & HG-104&- & 46.1 &30.0  & - &-  \\
     
     DETR-DC5 \cite{carion2020end} & ResNet-101 &  $800 \times 1333$ & 44.9 &23.7 & 60 & 254   \\

     RetinaNet \cite{zhang2021multi} & ViL-Small-RPB & $800 \times 1333$ &44.2 &28.8 & 35.68 & 254.8  \\

     YOLOv7-X \cite{wang2022yolov7}&-&$640\times640$ & 52.9 & - & 71.3 & 189.9 \\
     
     
     
     \bottomrule

    \end{tabular}
    
}
   \vspace{-2mm}
    \label{tab:validation_table}
\end{table}

\subsection{Results}

With no test-time augmentation, we compare YOGA with baselines at the image resolution of $640 \times 640$. Table \ref{tab:validation_table} reports the results on the validation dataset (5000 images with ground truth).  Table \ref{tab:testdev} reports the results on the test-dev dataset (20000 images with no public ground truth). In order to obtain the accuracy on test-dev, we submitted all our predictions to the {\tt\small CodaLab COCO Detection Challenge (Bounding Box)}~\cite{codalab} in JSON files. 
The AP\textsubscript{S}/AP\textsubscript{M}/AP\textsubscript{L} in Table \ref{tab:validation_table} and \ref{tab:testdev} means AP obtained on small/medium/large {\em objects} (not model scales). To simplify notation (e.g., in tables and figures), this section denotes by YOLO the YOLOv5 latest version v6.1 release in February 2022.


\blue{The scales of YOLO-v7 models (tiny-6.2M, base-36.9M, and X-71.3M) do not match our and and other baseline models, and thus prevent a fair comparison. For example, YOGA-\textit{n/m/l} has only 1.9/16.3/33.6 M parameters, hence we did not compare YOGA with YOLOv7. However, we still included YOLO-v7 results in both Tables \ref{tab:validation_table} and \ref{tab:testdev} for reference.}

\begin{table*}[t]
\caption{Results on {\tt MS-COCO \bf Test-Dev} Dataset. Percentages are in comparison against the {\bf closest} performer.}

\centering
\resizebox{\linewidth}{!}{%

\begin{tabular}{l|l|l|l|l|l|l|l|l|l|l} 
     \hline
     \hline
{\textbf{Method}} & {\textbf{ImgSize}} &{\textbf{FPS}}  & {\textbf{Params (M)}} & {\textbf{FLOPs (B)}} & {\textbf{AP[0.5: 0.95]}} & {\textbf{AP\textsubscript{50}}} &  {\textbf{AP\textsubscript{75}}}&  {\textbf{AP\textsubscript{S}}}  & {\textbf{AP\textsubscript{M}}}&  {\textbf{AP\textsubscript{L}}}\\ \midrule

YOGA-\textit{n} &640 &74 & 1.9 & 4.9 & \textbf{32.3} (\textcolor{blue}{\textbf{+15\%}}) & 50.3 &34.6&\textbf{14.2} (\textcolor{blue}{\textbf{+12\%}}) &\textbf{34.7} (\textcolor{blue}{\textbf{+11\%}})&\textbf{43.1} (\textcolor{blue}{\textbf{+22\%}}) \\
YOLO-\textit{n} \cite{yolov5} &640 & 158 & 1.9&4.5&28.1 & 45.7 & 29.8 & 12.7 & 31.3 & 35.4 \\
EfficientDet-D0 \cite{tan2020efficientdet} &512 &98.0 & 3.9&2.5& 33.8 & 52.2 & 35.8 & 12.0 & 38.3 & 51.2 \\
\midrule

YOGA-\textit{s} &640 & 67& 7.6 (\textcolor{red}{\text{+5\%}})& 16.6 (\textcolor{red}{\text{+0.6\%}})& \textbf{40.3} (\textcolor{blue}{\textbf{+8.6\%}}) & 59.1 &43.5&20.4&\textbf{43.5} (\textcolor{blue}{\textbf{+5\%}})&\textbf{53.1} (\textcolor{blue}{\textbf{+17\%}}) \\
YOLO-\textit{s} \cite{yolov5}&640 & 156 &7.2&16.5&37.1 & 55.7 & 40.2 & 20.0 & 41.5 & 45.2 \\
EfficientDet-D1 \cite{tan2020efficientdet} &640 &74.1& 6.6&6.1&39.6 & 58.6 & 42.3 & 17.9 & 44.3 & 56.0 \\
EfficientDet-D2 \cite{tan2020efficientdet} &768 &56.5& 8.1&11&43.0 & 62.3 & 46.2 & 22.5 & 47.0 & 58.4 \\
YOLOv7-tiny-SiLU \cite{wang2022yolov7} & 640 & 286 & 6.2 & 13.8 & 38.7 & 56.7 & 41.7 & 18.8 & 42.4 & 51.9 \\

\midrule

YOGA-\textit{m} &640 & 64& \textbf{16.3} (\textcolor{blue}{\textbf{-23\%}}) & \textbf{34.6} (\textcolor{blue}{\textbf{-29\%}}) & 46.4 & 65.0 & 50.3& 26.6& 50.1 & \textbf{58.9} (\textcolor{blue}{\textbf{+4\%}}) \\
YOLO-\textit{m} \cite{yolov5} &640 &121& 21.2&49.0& 45.5 & 64.0 & 49.7 & 26.6 & 50.0 & 56.6 \\
EfficientDet-D3 \cite{tan2020efficientdet} &896 &34.5& 12&25&45.8 & 65.0 & 49.3 & 26.6 & 49.4 & 59.8 \\
YOLOX-M \cite{ge2021yolox} &640 & 81.3& 25.3 & 51.4 & 46.4 & 65.4 & 50.6 & 26.3&51.0&59.9 \\
SSD512 \cite{liu2016ssd} &-&-&36.1&-&28.8&48.5&30.3&-&-&- \\
\midrule

YOGA-\textit{l} &640 & 62& \textbf{33.6} (\textcolor{blue}{\textbf{-27.7\%}}) & \textbf{71.8} (\textcolor{blue}{\textbf{-34\%}}) & 47.9 (\textcolor{red}{\text{-2.2\%}}) & 66.4 &51.9&28.0 (\textcolor{red}{\text{-6.3\%}})&51.6&60.6 \\
YOLO-\textit{l} \cite{yolov5} &640 &99& 46.5&109.1& 49.0 & 67.3 & 53.3 & 29.9 & 53.4 & 61.3 \\
YOLOX-L \cite{ge2021yolox} &640 &69.0 &54.2 &115.6& 50.0 & 68.5 & 54.5 & 29.8 &54.5 &64.4 \\
YOLOv4-CSP \cite{wang2021scaled} &640& 73&52.9&-& 47.5&66.2&51.7&28.2&51.2&59.8 \\
PP-YOLO \cite{long2020pp} &608&72.9&52.9&-& 45.2& 65.2& 49.9& 38.4& 59.4&67.7 \\
YOLOv7 \cite{wang2022yolov7} & 640 & 161 & 36.9 & 104.7 & 51.4 & 69.7 & 55.9 & 31.8 & 55.5 & 65.0 \\
 \midrule



YOLOX-X \cite{ge2021yolox} &640 & 57.3& 99.1 & 219.0 & 51.2 & 69.6 & 55.7 & 31.2  &56.1 &66.1\\ 
YOLOv4-P5 \cite{wang2021scaled} &896&43&70.8&-& 51.8&70.3&56.6&33.4&55.7&63.4 \\
YOLOv4-P6 \cite{wang2021scaled} &1280&32&127.59&-& 54.5& 72.6& 59.8& 36.8& 58.3&65.9\\
YOLOv4-P7 \cite{wang2021scaled} &1536&17&287.57&-&55.5& 73.4& 60.8&38.4& 59.4&67.7 \\ 
EfficientDet-D5 \cite{tan2020efficientdet}  & 1280 & - & 34 & 135 & 51.5 & 70.5 & 56.7 & 33.9 & 54.7 & 64.1 \\
ATSS & 800 &- & - & - & 46.3 & 64.7 & 50.4 & 27.7 & 49.8 & 58.4  \\
EfficientDet-D6 \cite{tan2020efficientdet} & 1280 & -&52 & 226 & 52.6 & 71.5 & 57.2 & 34.9 & 56.0 & 65.4 \\
RDSNet (R-101) \cite{wang2020rdsnet} & 800 &-& 0 & - & 38.1 & 58.5 & 40.8 & 21.2 & 41.5 & 48.2 \\
RetinaNet (SpineNet-143) & 1280 & -& 66.9 & 524.4 & 50.7 & 70.4 & 54.9 & 33.6 & 53.9 & 62.1 \\

YOLOv7-X \cite{wang2022yolov7} & 640 & 114 & 71.3 & 189.9 & 53.1 & 71.2 & 57.8 & 33.8 & 57.1 & 67.4 \\


\bottomrule

\end{tabular}

\label{tab:testdev}
}

\end{table*}



\pgfdeclareplotmark{mystar}{
    \node[star,star point ratio=2.25,minimum size=10pt,
    inner sep=0pt,draw=red,solid,fill=red] {};
}

\pgfdeclareplotmark{nabla}{
    \node[nabla,star point ratio=2.25,minimum size=5pt,
    inner sep=0pt,draw=orange,solid,fill=orange] {};
}


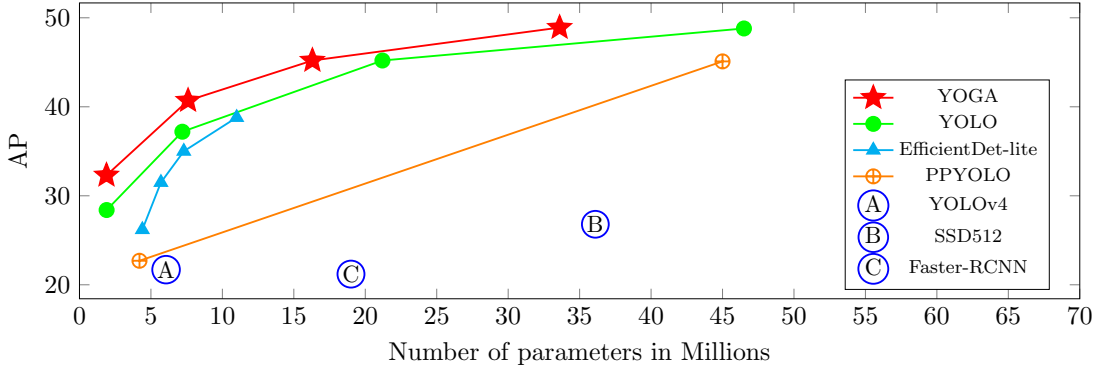
\begin{figure}[ht]

\centering

\resizebox{0.9\textwidth}{!}{
\begin{tikzpicture}[][x=0.5pt,y=5pt,yscale=-1,xscale=1]

\begin{axis}[
	xlabel= \text{Number of parameters in Millions},
	ylabel= \text{AP},
	width= \linewidth,height=6cm,
	ylabel near ticks,
	xmin=0, xmax=70,
    legend pos= south east, legend style={font=\scriptsize}]

\addplot[color=red, mark=mystar, thick, mark size=3pt] plot coordinates {
    (1.89, 32.3)
	(7.6, 40.7)
	(16.3, 45.2)
	(33.6, 48.9)
};

\addplot[thick,color=green,mark=*,mark size =3pt] plot coordinates {
     (1.9,28.4)
     (7.2,37.2)
     (21.2,45.2)
     (46.5, 48.8)
     };
     
\addplot[mark=triangle*, color=cyan, mark size=3pt, thick] plot coordinates {
	(4.4, 26.2)
	(5.7, 31.5)
	(7.3, 35)
	(11, 38.8)};

\addplot[mark=oplus,mark size=3pt,draw=orange,thick] plot coordinates {
   (4.20,22.7)
   (45,45.1)
};


\addplot[only marks,
      mark=text,
      text mark=A,
      text mark as node,
      text mark style={%
        font=\small,
        circle,
        inner sep=.05em,
        fill=blue!10!white,
        draw=blue,thick,
        fill=none
      }] plot coordinates {
 (6.06,21.7)
};

\addplot[only marks,
      mark=text,
      text mark=B,
      text mark as node,
      text mark style={%
        font=\small,
        circle,
        inner sep=.05em,
        fill=blue!10!white,
        draw=blue,thick,fill=none
      }] plot coordinates {
 (36.1,26.8)
};

\addplot[only marks,
      mark=text,
      text mark=C,
      text mark as node,
      text mark style={%
        font=\small,
        circle,
        inner sep=.05em,
        fill=blue!10!white,
        draw=blue,thick, fill=none
      }] plot coordinates {
(19, 21.2)
};


\legend{YOGA, YOLO,EfficientDet-lite, PPYOLO,YOLOv4, SSD512, Faster-RCNN}
\end{axis}

\end{tikzpicture}
}
\caption{Comparing YOGA with state-of-the-art object detection models. 
}\vspace{-1mm}
\label{figure:teaser}
\end{figure}


\fref{figure:teaser} provides a comparison of YOGA with multiple SOTA models in terms of AP and number of parameters. The four YOGA points correspond to YOGA-\textit{n}/\textit{s}/\textit{m}/\textit{l}. Similarly, the points of other models correspond to their respective model sizes too. The results show that YOGA has the best AP at every model scale, or equivalently the lightest model for any target AP. For instance, PPYOLO ~\cite{long2020pp} has an AP of 22.7 at 4.20M parameters, while YOGA achieves an AP of $\sim$35 (interpolated) with the same number of parameters, amounting to a 54\% improvement. More thorough and detailed comparisons are presented in Tables \ref{tab:validation_table} and \ref{tab:testdev} and discussed below.

\textbf{Nano and Small models.} With the same number (1.9M) of parameters, YOGA-\textit{n} achieves an AP of 32.3 which is 15.35\% higher than the best-performing model, YOLO-\textit{n}, whose AP is 28.0. In the comparison on small objects, YOGA-\textit{n} achieves an improvement of 7.4\% AP\textsubscript{S} over YOLO-\textit{n}.

Similarly, our YOGA-\textit{s} achieves 40.7 AP while the best-performing YOLO-\textit{s} achieves 37.4, with almost the same number of parameters and FLOPs, which indicates a 8.8\% increase in the AP value. Our YOGA-\textit{s} achieves 23.0 AP on small objects (AP\textsubscript{S}) while the best-performing YOLO-\textit{s} achieves 21.09 AP\textsubscript{S}, a +9.0\% increase in the AP\textsubscript{S}.
Compared to state-of-the-art models on test-dev (Table \ref{tab:testdev}), our YOGA-\textit{n} model compared to YOLO-\textit{n} achieves an improvement of 15\% AP (+4.2 AP). When we compare on object scales, on small objects, there is an improvement of 12\% AP\textsubscript S (+1.5 AP\textsubscript{S});  
on medium objects, there is an improvement of 11\% AP\textsubscript{M} (+3.4 AP\textsubscript{M} ), and on large objects, there is an improvement of 22\% AP\textsubscript{L} (+7.7 AP\textsubscript{L} ).


Similarly, our YOGA-\textit{s} model compared to YOLO-\textit{s}, achieves an improvement of 8.6\% AP (+3.2 AP). When we compare on object scales, on medium objects, there is an improvement of 5\% AP\textsubscript{M} (+2.0 AP\textsubscript{M});  
on large objects, there is an improvement of 17\% AP\textsubscript{L} (+7.9 AP\textsubscript{L}).  

\textbf{Medium and Large models.} When compared to YOLO-\textit{m}, our YOGA-\textit{m} achieves similar AP value of 45.2 but significantly reduces parameters and FLOPs by 23\% and 29\% respectively. simlarly, comparing with YOLOX-M model, our YOGA-\textit{m} has significantly reduction in parameters and FLOPs by 35\% and 53\%.

Compared to YOLO-\textit{l}, our YOGA-\textit{l} model achieves the same AP (48.9) but with a significantly lower number of parameters and FLOPs: 27.7\% and 34\% respectively. simlarly, comparing with YOLOX-L, our YOGA-\textit{l} model has significant reduction in parameters and FLOPs by 38\% and 53.8\%. 

When compared to state-of-the-art models on test-dev (Table \ref{tab:testdev}), YOGA-\textit{m} achieves AP of 46.4, a 2\% improvement to YOLO-\textit{m} and it uses only 16.3 M parameters and and 34.6 BFLOPs, which are significantly (23\% and 29\%) lower than YOLO-\textit{m}.

    



\begin{figure*}[t]
    \centering
    \subfloat[Green boxes: YOLO-\textit{n} predictions.]{\includegraphics[width=0.47\linewidth]{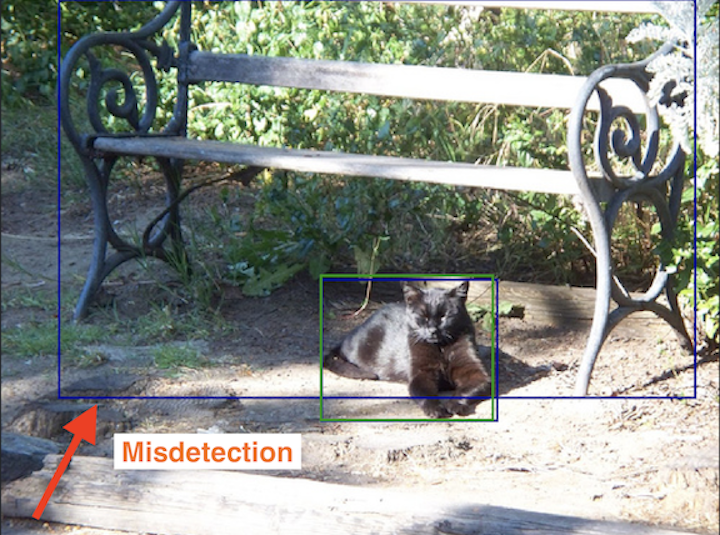} }
    \subfloat[Purple boxes: YOGA-\textit{n} predictions.]{\includegraphics[width=0.47\linewidth]{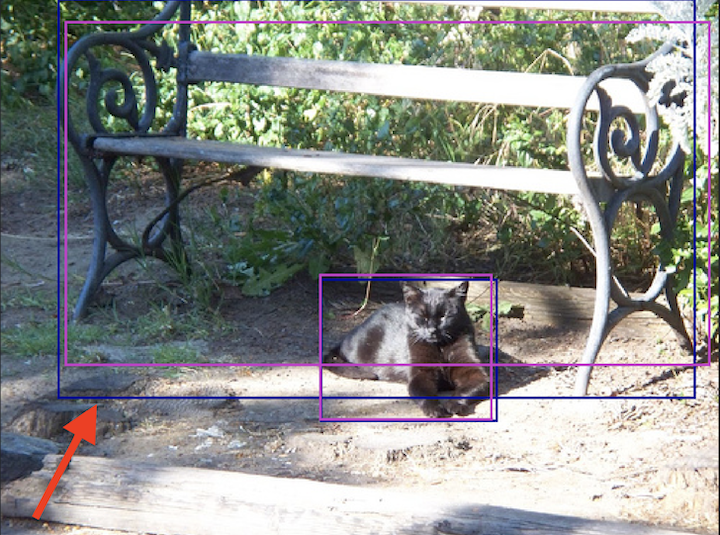}}
    \\
    \subfloat[Green boxes: YOLO-\textit{n} predictions.]{\includegraphics[width=0.47\linewidth]{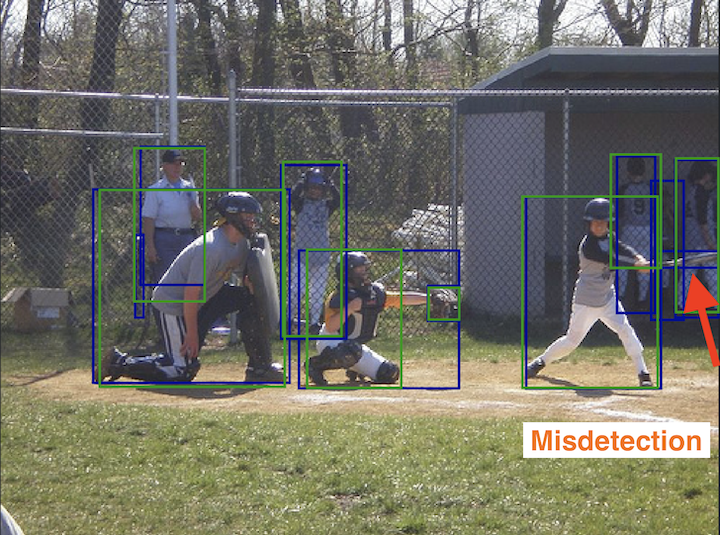} }
    \subfloat[Purple boxes: YOGA-\textit{n} predictions.]{\includegraphics[width=0.47\linewidth]{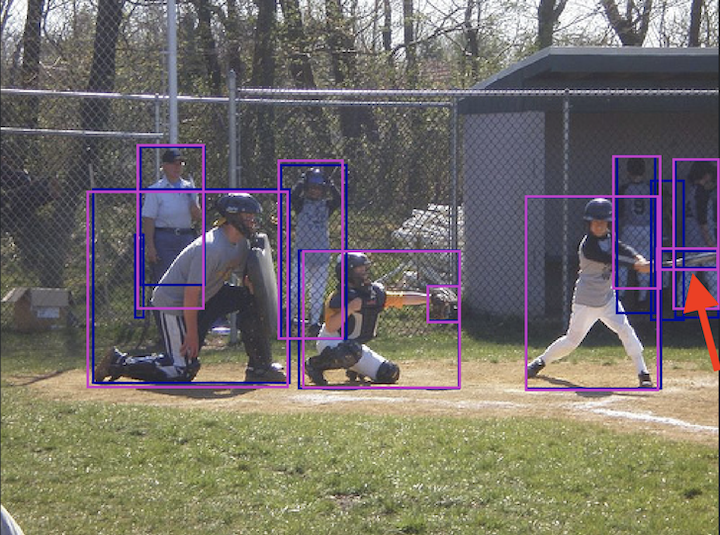} }

    \caption{A visual comparison. Blue boxes: the COCO-17 ground truth. Red arrows highlight the differences.}
    \label{fig:object_comparison}
\end{figure*}

{\bf Visual Comparison.}
We compare the bounding box predictions of the YOLO-\textit{n} and our YOGA-\textit{n} model on two random sample images on COCO validation dataset. In \fref{fig:object_comparison}, we see that the top row image has two ground truth objects (Blue boxes) and our YOGA-\textit{n} detects both objects (Purple boxes), while YOLO-\textit{n} (Green boxes) fail to detect one object (Bench). Similarly, the bottom row image has a total of eleven ground truth objects, and YOLO-\textit{n} detects only eight objects while our YOGA-\textit{n} model detects nine objects, an extra object called Baseball bat.

\subsection{Hardware Implementation and Evaluation}


To assess the edge suitability of YOGA and its usability in the wild, we migrate YOGA code to NVIDIA Jetson Nano 2GB, which  comes with 2 GB 64-bit LPDDR4 25.6 GB/s RAM and 32 GB MicroSD storage 
and is the {\em lowest-end} deep learning hardware product from NVIDIA.

\begin{figure}
    \centering
    \includegraphics[width=\linewidth]{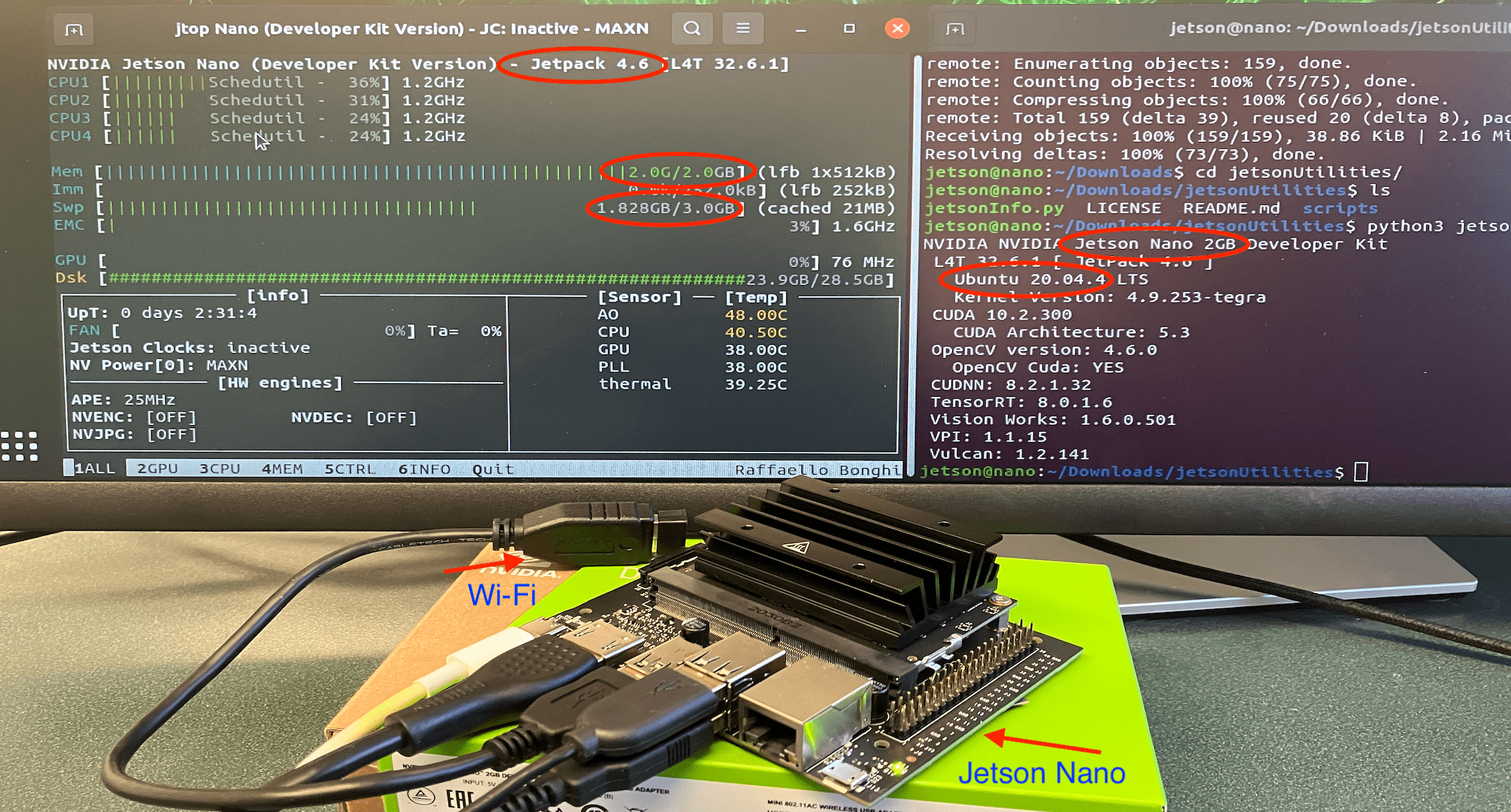}
    \caption{Our hardware testbed setup and run-time outputs.}
    \label{fig:hardware_setup}
\end{figure}

\fref{fig:hardware_setup} shows the hardware setup for our edge inference experiments, where we have set up the runtime environment (Ubuntu 20.04, PyTorch 1.12, Jetpack 4.6) for evaluation. We measure the inference time of YOGA to see how near-real-time it can be when performing object detection. The results are reported in \tref{tab:nano_inference}, where we see that YOGA-n achieves an inference time of of 0.57 sec per image (each image is of large size 640x640) which is close to real-time. We highlight an important fact that, as seen from \fref{fig:hardware_setup} (in oval shapes), the 2GB memory on Jetson Nano was fully utilized at peak time and 1.828 GB swap space on the disk had to be used to compensate for the memory shortage. This means that the disk I/O had throttled the performance substantially, and it is therefore reasonable to anticipate a significantly better performance on the 4 GB Jetson Nano and even better on Jetson TX2, which would no longer or rarely need to use swap space, making the inference indeed (near)real-time.\footnote{Both before and at the time of writing, the global market has been undergoing a severe GPU product shortage and many products have been out of stock in the market. As a consequence, we could not procure more hardware for testing.}

\begin{table}[!h]
\caption{Performance on Jetson Nano 2GB with 640 x 640 (large) COCO images.}\vspace{-1mm}
    \centering
\resizebox{.37\linewidth}{!}{%
    \begin{tabular}{lc}
     \hline
     {Model} & {Inference time (sec)}  \\ \midrule
      YOGA-\textit{n} & 0.57   \\
      YOGA-\textit{s} & 0.77   \\
      YOGA-\textit{m} & 0.98  \\
      YOGA-\textit{l} &1.30 \\
    \bottomrule
    \end{tabular}
     
    }
    \label{tab:nano_inference}
\end{table}

\subsection{Ablation study}
We also design experiments to investigate the individual effect of our new backbone and neck: specifically, how our CSPGhostNet backbone compares to the YOLO backbone (arguably the best backbone so far) and how our AFF-PANet neck compares to the naive concatenation as used in all SOTA architectures. Moreover, we also evaluate the effect of label smoothing on the gradient descent convergence. We conduct these ablation studies using YOGA-\textit{n}.

\begin{table}[h]
\caption{Ablation study on Backbone and Neck (YOGA-{\it n}).}
    \centering
    \small
\resizebox{.45\linewidth}{!}{
    \begin{tabular}{ccc}
     \hline
     
     {Backbone} & {Neck} &{AP} \\ \midrule
     
     YOLO Backbone & Naive Concat & 28.4 \\
     \blue{YOLO Backbone} & \blue{AFF-PANet} & \blue{29.2} \\
     CSPGhostNet & Naive Concat & 31.1\\
     CSPGhostNet & AFF-PANet &  \textbf{32.3}\\
     \bottomrule
    \end{tabular}
}
\label{table:4}
\vspace{-1mm}
\end{table}

%
%
%
    


\blue{The results for backbone and neck are given in Table \ref{table:4}. We observe that, using the AFF-PANet neck architecture consistently leads to improved performance compared to using the Naive Concat (PANet) neck architecture. Additionally, using our CSPGhostNet backbone leads to better performance than using the existing YOLO backbone in both cases. Overall, these results suggest that both the AFF-PANet neck and the CSPGhostNet backbone contribute positively to the performance of YOGA.}


For label smoothing, we observed during our training that it helped our model training to converge to a desirable AP[0.5:0.95] and recall in $\sim$10\% less number of epochs than without label smoothing.



\section{Conclusion}

\blue{This paper presents YOGA, a novel object detection model with an efficient convolutional backbone and an enhanced attention-based neck. It is a deep yet lightweight object detector with high accuracy, which we have validated with extensive evaluation benchmarked against more than 10 state-of-the-art modern deep detectors. For instance, in its Nano version, our YOGA-\textit{n} outperforms the current best-performing model YOLOv5n by 15.35\% in AP, with similar number of parameters and FLOPs; this improvement further increases to 22\% on detecting large objects (AP\textsubscript{L}) on the test-dev dataset. In its Medium version, our YOGA-\textit{m} achieves the same AP (45.2) as the best-performing model YOLOv5m but with 23\% fewer parameters and 29\% fewer FLOPs. In its Large version, our YOGA-\textit{l} achieves the same AP (48.9) as the best-performing model YOLOv5-l but with 27.7\% fewer parameters and 34\% fewer FLOPs.}

\blue{We have also implemented and assessed YOGA on the {\em lowest-end} deep learning device from NVIDIA, Jetson Nano 2GB, and the results affirmed that YOGA is suitable for edge deployment in the wild. For instance, YOGA-\textit{n} runs at 0.57 sec per 640x640 image, which is close to real-time.}

\blue{The main limitation of YOGA is that it could be prone to overfitting when training extremely large models. Nonetheless, such extra-large models are rather unlikely to be adopted in edge deployments. Future directions for improving YOGA or object detection in general, include: (1) investigating different attention mechanisms such as incorporating self-attention or transformer architectures; (2) exploring ways to further optimize the model for specific hardware platforms such as mobile devices; (3) extending YOGA to handle additional tasks or challenges such as semantic segmentation, instance segmentation, and object tracking.}

\blue{In summary, YOGA represents a new contribution to the field of object detection by ushering in high run-time efficiency, low memory footprint, and superior accuracy simultaneously. In addition, its flexible scalability makes it applicable to a wide range of applications with different hardware constraints in IoT, edge and cloud computing.}



\bibliographystyle{elsarticle-harv} 
\bibliography{ref}





\end{document}